\documentclass{article}

 \usepackage[dblblindworkshop, final, nonatbib]{neurips_2026}
 
\usepackage[utf8]{inputenc} 
\usepackage[T1]{fontenc}    
\usepackage{hyperref}       
\usepackage{url}            
\usepackage{booktabs}       
\usepackage{amsfonts}       
\usepackage{nicefrac}       
\usepackage{microtype}      
\usepackage{xcolor}         
\usepackage{pgfplots}

\usepackage[
  backend=biber,
  natbib,
  style=numeric-comp,
  maxbibnames=99,
  sorting=none,
  alldates=year,
  isbn=false,
  doi=true,
]{biblatex}

\title{Non-Interfering Asymmetric AdamW Updates\\ for Constrained Exploration of Equifinality\\ in Hybrid Models}

\workshoptitle{Sim2Science: ML with Imperfect Scientific Models}

\author{
    Ekaterina Gikalo\textsuperscript{1,2} \\
    \texttt{egikalo@bgc-jena.mpg.de}
    \And
    Zavud Baghirov\textsuperscript{1,2} \\
    \texttt{zbaghirov@bgc-jena.mpg.de}
    \And
    Christian Reimers\textsuperscript{1,3} \\
    \texttt{creimers@bgc-jena.mpg.de} 
    \And
    Markus Reichstein\textsuperscript{1,3} \\
    \texttt{mreichstein@bgc-jena.mpg.de}
    \And
    Martin Jung\textsuperscript{1} \\
    \texttt{mjung@bgc-jena.mpg.de}
    \And
    Marco Körner\textsuperscript{4,2,3} \\
    \texttt{marco.koerner@dlr.de}
    \And
    \normalfont
    \textsuperscript{1}Max Planck Institute for Biogeochemistry, Jena, Germany \\
    \textsuperscript{2}TUM School of Engineering and Design, Technical University of Munich (TUM), Munich, Germany \\
    \textsuperscript{3}ELLIS Unit Jena, Jena, Germany \\
    \textsuperscript{4}German Remote Sensing Data Center (DFD), German Aerospace Center (DLR), Weßling, Germany\\
}

\newcommand{\Lmain}{\mathcal{L}_{\mathrm{main}}}
\newcommand{\Laux}{\mathcal{L}_{\mathrm{aux}}}
\newcommand{\gmain}{\V{g}_{\mathrm{main}}}
\newcommand{\gaux}{\V{g}_{\mathrm{aux}}}

\newcommand{\gtil}{\tilde{\V{g}}_{\mathrm{aux}}}
\newcommand{\mmain}{\V{m}_{\mathrm{main}}}

\newcommand{\mhatmain}{\hat{\V{m}}_{\mathrm{main}}}

\newcommand{\vmain}{\V{v}_{\mathrm{main}}}
\newcommand{\vhatmain}{\hat{\V{v}}_{\mathrm{main}}}

\newcommand{\Pmain}{\M{P}_{\mathrm{main}}}
\newcommand{\Paux}{\M{P}_{\mathrm{aux}}}
\newcommand{\qmain}{q_{\mathrm{main}}}
\newcommand{\qaux}{q_{\mathrm{aux}}}
\newcommand{\cmain}{c_{\mathrm{main}}}
\newcommand{\caux}{c_{\mathrm{aux}}}
\newcommand{\inner}[2]{\langle #1, #2 \rangle}
\newcommand{\dsum}{\V{d}_{\mathrm{sum}}}
\newcommand{\dpc}{\V{d}_{\mathrm{PC}}}

\usepackage{amsmath}
\usepackage{amsthm}
\usepackage{algorithm}
\usepackage{algpseudocode}

\usepackage{tummath}
\usepackage{tumabbrev}
\usepackage[inline]{enumitem}

\usepackage{zref-clever}
\zcsetup{cap}

\usepackage[
  acronym,
  automake,
]{glossaries-extra}
\makeglossaries

\setabbreviationstyle[acronym]{long-short-user}
\renewcommand*{\glsxtruserparen}[2]{
  \glsxtrfullsep{#2}%
  \glsxtrparen
   {#1\ifglshasfield{\glsxtruserfield}{#2}{;\xspace%
     \expandafter\citealp\expandafter{\glscurrentfieldvalue}%
   }{}%
   }%
}

\glsdefpostdesc{acronym}{
 \ifglshasfield{\glsxtruserfield}{\glscurrententrylabel}%
 {~\expandafter\citep\expandafter{\glscurrentfieldvalue}}%
 {}%
}

\glsdisablehyper
\defglsentryfmt[acronym]{\ifglsused{\glslabel}{\glsgenentryfmt}{\emph{\glsgenentryfmt}}}

\newacronym{gl:H2CM}{H2CM}{Hybrid Hydrological Carbon Cycle Model}
\newacronym{gl:PCGrad}{PCGrad}{Projecting Conflicting Gradients}
\newacronym{gl:Bloop}{Bloop}{BiLevel Optimisation with Orthogonal Projection}
\newacronym{gl:NANO}{NANO}{Non-interfering Asymmetric Network Optimisation}

\begin{document}

\newtheorem{proposition}{Proposition}
\newtheorem{lemma}{Lemma}
\theoremstyle{remark}
\newtheorem{remark}{Remark}

\maketitle

\begin{abstract}
Hybrid models let neural networks supply the unresolved parameters of a process-based simulator, and are fitted to observations that constrain aggregate fluxes and storages but not the internal partitioning that produces them.
Such a fit is \emph{equifinal}: configurations indistinguishable in the observed aggregates can disagree substantially about the unobserved internals, and therefore about the mechanism they attribute to the same data.
We study evapotranspiration, whose total flux is observed while its split into transpiration and evaporation is not.
A model ensemble built from random seeds alone spans only a narrow band of the partitionings the data admit.
We produce the spread deliberately, training members under auxiliary repulsion on transpiration, subject to the fit to observations not degrading.
Non-interference is usually arranged by making the auxiliary gradient orthogonal to the main one, but under AdamW the applied step is a preconditioned momentum direction, so orthogonality between raw gradients no longer implies it.
We introduce \gls{gl:NANO}, an update whose first-order effect on the main loss is exactly that of the main objective alone, whose auxiliary step is bounded, and which concentrates repulsion in the coordinates the main gradient leaves idle.
\end{abstract}

\section{Problem setting}
\label{sec:problem-setting} 

Hybrid models embed neural networks in a process-based simulator and train them end-to-end through differentiable physics, so that known structure is retained and the unresolved parameters are fitted to data \citep{reichstein2019}. 
A model is \emph{equifinal} when distinct configurations of it reproduce the available observations equally well, so that the data select no single mechanism among them \citep{beven2006}.
In a hybrid model, equifinality can arise from several sources at once: different neural network outputs can produce the same simulated targets; the process model may be too simplified to describe the full system; its components may be interchangeable, a deficit in one being compensated by an excess in another; and the observations could constrain only a few aggregate dimensions of a much higher-dimensional internal state.

\begin{figure}[t]
    \centering
    \includegraphics[width=0.9\linewidth, height=7cm, keepaspectratio]{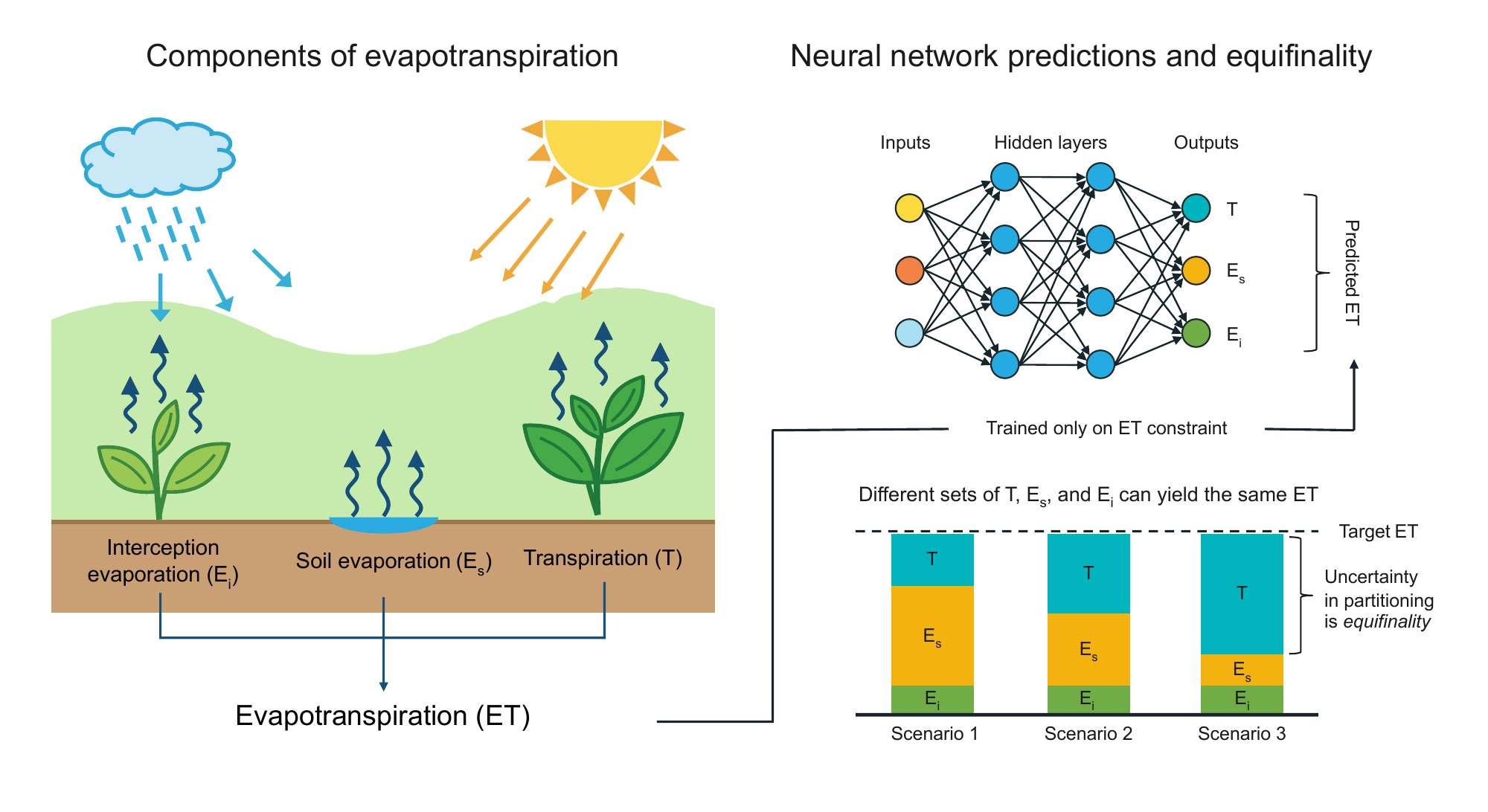}
    \caption{Equifinality in the partitioning of evapotranspiration.
    Left: $\mathrm{ET}$ is the sum of transpiration, soil evaporation and interception evaporation.
    Right: network is trained against $\mathrm{ET}$ alone, so different partitionings can reproduce the same observed target.}
    \label{fig:et}
\end{figure}

Our study focuses on \emph{evapotranspiration}, \(\mathrm{ET} = T + E_s + E_i\), whereby water is transported from the Earth's surface into the atmosphere through \emph{evaporation}---\ie soil evaporation $E_s$ from the soil surface and interception evaporation $E_i$ from canopies before it reaches the ground---, as well as through stomatal \emph{transpiration} $T$ from vegetation stocks (\cf \zcref{fig:et}). 
%
%
While the transpiration fraction \(R = T/\mathrm{ET}\) is key to understanding the water-carbon cycle, the coupling between terrestrial water fluxes and carbon uptake, and the response to drought and irrigation, only the total $\mathrm{ET}$ is constrained through observations.
%
Published global estimates of $R$ vary widely, between \numrange[round-mode=places,round-precision=2]{.8}{.9} \citep{jasechko2013} and \num[separate-uncertainty]{0.64 \pm 0.13} \citep{good2015}.
Making this equifinality explicit therefore allows a model to report honestly what the observations do and do not determine.

The \gls{gl:H2CM} \citep{baghirov2026} is a hybrid model of the coupled water and carbon cycles: recurrent and feed-forward networks predict its process parameters, the physics propagates them into states and fluxes, and these are compared against spatio-temporal observations.
While $\mathrm{ET}$ is among those targets, listed in \zcref{app:h2cm_loss}, its partitioning is not.

\section{Methodology}
\label{sec:methodology} 
Training a standard random-seed ensemble underestimates the equifinality in \gls{gl:H2CM}, as no term in the main objective distinguishes members that agree on $\mathrm{ET}$ but disagree on its partitioning.
Instead, we produce the spread deliberately: $M$ members are trained under an auxiliary repulsion loss $\Laux$. 
The difficulty lies in adding the repulsion without increasing the main loss $\Lmain$.

Multi-task methods rebalance or reshape task gradients \citep{yu2020, sener2018, liu2021, hsieh2024, Liebel18:ATM, farajtabar2019}, while asymmetric variants protect a designated main task by gating or decomposing the auxiliary gradient \citep{du2018, dery2021}. 
Closest to our setting are \gls{gl:PCGrad} \citep{yu2020}, which introduces orthogonal projection between two gradients, and \gls{gl:Bloop} \citep{hsieh2024}, which projects the auxiliary gradient orthogonally to an exponential moving average of the main gradient.
All of them, however, are formulated for raw gradients under plain descent.

Under Adam and AdamW \citep{kingma2015, loshchilov2019} the applied step is a diagonally preconditioned momentum direction, so orthogonality between raw gradients no longer implies non-interference, and moments accumulated from a combined direction carry the auxiliary signal into the main task's steps.
Methods that adapt the optimiser itself keep per-task or per-term moments \citep{yang2023, malkiel2021}, or, in continual learning, project a new task's Adam update into the null space of previous tasks' features \citep{wang2021}, protecting tasks that are no longer trained.
We instead protect the main task trained concurrently with the auxiliary one, and keep both moments free of the auxiliary signal, which requires an update that

\newlist{critlist}{enumerate}{1}%
\setlist[critlist]{font=\bfseries, label=D\arabic*, ref=D\arabic*}\zcRefTypeSetup{critlisti}{%
  name-sg = ,%
  Name-sg = ,%
  name-pl = ,%
  Name-pl = ,%
}%
\begin{critlist}[]
  \item \label{update:fo-descent}
  leaves the descent on the main objective unchanged to first order, equal to what the training on the main task would have produced alone;
  
  \item \label{update:so-bounded}
  bounds the loss the auxiliary term adds to the main objective to second order;
  
  \item \label{update:aux-biased}
  biases the auxiliary term toward coordinates the main gradient does not persistently use.   
  
\end{critlist}

The contributions of this paper are:
\begin{enumerate*}[label=\textit{\roman*)}, itemjoin={,\xspace}, itemjoin*={{, and\xspace}}]
  \item we formulate the search for equifinal diversity as a constrained problem and derive a cheap local surrogate for it
  \item we introduce \gls{gl:NANO}, an optimiser-level construction that realises \zcref[rangesep={--}]{update:fo-descent,update:so-bounded,update:aux-biased} within AdamW
  \item we show on \gls{gl:H2CM} that it produces spread in the unobserved transpiration fraction $R=T/\mathrm{ET}$, without degrading the fit to the observations.
\end{enumerate*}

\subsection{Repulsion loss}
\label{sec:repulsion}

We train an ensemble of $M$ models on identical batch sequences.
Let \(\V{\theta}^{(m)} \in \mathbb{R}^{n}\) be the parameters of one member $m$, $\Lmain$ its loss to the observed targets, and \(\mu(\V{\theta}^{(m)}) \in \mathbb{R}\) an internal quantity of the hybrid model, not constrained directly by those targets.
In our experiments, \(\mu(\V{\theta}^{(m)})\) is the spatio-temporal mean transpiration $T$.
With \(N(m) = \{\nu^{-}, \nu^{+}\}\) we denote the closest diagnostic values among the other members below and above \(\mu(\V{\theta}^{(m)})\). 
The repulsion loss is
\begin{align}
  \Laux\left(\V{\theta}^{(m)}\right) &= \tau \operatornamewithlimits{mean}_{\nu \in N(m)} \exp\!\left(- \frac{r_{\nu}}{\tau}\right), \qquad r_{\nu} = \sqrt{(\mu(\V{\theta}^{(m)}) - \nu)^{2} + \varepsilon_{\mu}^{2}},
\label{eq:repulsion}
\end{align}
where $\varepsilon_{\mu} > 0$ is a small constant for numerical stability and $\tau > 0$ sets the range of the repulsion between members.
Since $\nu^{\pm}$ move as the other members train, $\Laux$ is \emph{non-stationary} even at fixed \(\V{\theta}^{(m)}\) and the scheme is a simultaneous update rather than gradient descent on a joint objective.

What we would like to find is
\begin{align}
\min_{\V{\theta}^{(1)},\dots,\V{\theta}^{(M)}}\ \sum_{m} \Laux\big(\V{\theta}^{(m)}\big) \qquad\text{s.t.}\qquad \Lmain\big(\V{\theta}^{(m)}\big) \le \Lmain^{\star} + \delta \quad \forall 1 \leq m \leq M,
\label{eq:constrained}
\end{align}
where $\Lmain^{\star}$ is the loss reachable without the auxiliary term and $\delta$ is a small constant.
Note that we do not solve \zcref{eq:constrained} directly, we only use it to motivate the local surrogate derived next.

\subsection{Non-interfering updates under AdamW}
\label{sec:method}

\paragraph{Why a weighted sum fails.}
At step $t$ with learning rate $\alpha$, we define both gradients as \(\gmain = \nabla_{\V{\theta}} \Lmain(\V{\theta}^{(m)}_{t})\) and \(\gaux = \nabla_{\V{\theta}} \Laux(\V{\theta}^{(m)}_{t})\) for a model $m$.
The default multi-objective construction with a weight $w$ is $\mathcal{L} = \Lmain + w\,\Laux$ and gives the direction $\dsum = \gmain + w\,\gaux$.
This violates \zcref{update:fo-descent}, as the auxiliary term contributes $-\alpha w \inner{\gmain}{\gaux}$ to $\Delta \Lmain$, \ie an increase in the main loss whenever $\inner{\gmain}{\gaux} < 0$.
It also offers no stable scale for $w$, as $\|\gaux\|$ is set by the inter-member distances and decays as members separate.
The one-sided adaptation of \gls{gl:PCGrad} to our asymmetric setting projects the auxiliary gradient only and leaves $\gmain$ untouched. 
With the projection \(\pi(\V{x};\V{y}) = \V{x} - \frac{\inner{\V{x}}{\V{y}}}{\|\V{y}\|^{2}}\,\V{y}\), the update direction becomes $\dpc = \gmain + \pi(\gaux;\gmain)$, for which $\inner{\gmain}{\dpc} = \|\gmain\|^{2}$ holds.
Hence, to first order, the auxiliary term moves $\V{\theta}$ only along directions in which $\Lmain$ is constant.
This realises \zcref{update:fo-descent} for plain gradient descent, but it does not carry over to Adam optimiser's preconditioning.
 
\paragraph{Keeping the moments free of the auxiliary signal.}
Adam optimisation applies a step $-\alpha \M{P} \hat{\V{m}}$, where $\hat{\V{m}}$ is a bias-corrected first moment and $\M{P} = \operatorname{diag}\big((\sqrt{\hat{\V{v}}})^{-1}\big)$ is constructed from the second moment. 
Running Adam on the summed objective accumulates \(\V{m} \leftarrow \beta_{1} \V{m} + (1-\beta_{1})(\gmain + \gaux)\) and \(\V{v} \leftarrow \beta_{2} \V{v} + (1-\beta_{2})(\gmain + \gaux)^{2}\). 
Since $\M{P}$ depends on $\gaux$, the main task's per-coordinate step sizes are affected by the auxiliary signal and inherit its non-stationarity.
We therefore accumulate both moments from the main gradient alone, \ie
\begin{align}
\mmain \leftarrow \beta_{1} \mmain + (1-\beta_{1}) \gmain, \qquad \vmain \leftarrow \beta_{2} \vmain + (1-\beta_{2})\, \gmain^{2},
\label{eq:moments}
\end{align}
with the usual bias corrections, and precondition main and auxiliary tasks with $\vhatmain$.
This is how \zcref{update:aux-biased} is realised: the auxiliary term is scaled by the coordinate-wise history of the main gradient, damped where the main objective is persistently active, and amplified where it is idle.
That amplification is unbounded, as \(\sqrt{\vhatmain^{(i)}} \to 0\), so for a floor fraction $c \ge 0$ we define
\begin{align}
\M{P}_{c} = \operatorname{diag}\!\Big( \big( \sqrt{\vhatmain} + \V{\varepsilon}_{c} \big)^{-1} \Big), \qquad \varepsilon_{c}^{(i)} = \max\!\Big( c \cdot \operatornamewithlimits{median}_{j \in B(i)} \sqrt{\vhatmain^{(j)}},\; \varepsilon_{0} \Big),
\label{eq:precond}
\end{align}
where $\varepsilon_{0}$ is a small constant for numerical stability and \(B(i)\) is the parameter tensor containing coordinate $i$, so the floor is relative to that tensor's own scale.
We use $\Pmain = \M{P}_{\cmain}$ and $\Paux = \M{P}_{\caux}$ with $\caux \ge \cmain$, so the floor bounds the auxiliary amplification where the main gradient is idle, while $\cmain = 0$ leaves the main task optimisation exactly AdamW's.
 
\paragraph{Projecting after preconditioning.}
An exponential moving average estimates a quantity assumed fixed on the averaging timescale, whereas the auxiliary signal moves with the other members, so we use the instantaneous gradient $\gaux$.
Prior to any correction, the step direction is $\Pmain \mhatmain + \Paux \gaux$, for which
\begin{align}
  \Delta \Lmain &\approx -\alpha \inner{\gmain}{\Pmain \mhatmain} - \alpha \inner{\gmain}{\Paux \gaux},
  \label{eq:firstorder}
\end{align}
so \zcref{update:fo-descent} requires $\inner{\gmain}{\Paux \gaux} = 0$.
The preconditioned auxiliary direction, not the raw gradient, must be orthogonal to the main gradient.
We also remove a second direction $\mhatmain$ to bound the second-order term and satisfy \zcref{update:so-bounded}, as described subsequently.

The projection \(\Pi(\V{x};\V{y},\V{z}) = \pi\big(\pi(\V{x};\V{y});\, \pi(\V{z};\V{y})\big)\) is constructed by applying the projection $\pi$ twice, first against $\V{y}$ and then against the part of $\V{z}$ that is orthogonal to $\V{y}$, resulting in a projection
\begin{align}
  \gtil = \Pi\big(\Paux \gaux;\, \gmain,\, \mhatmain\big)
  \label{eq:projpost}
\end{align} against the span of $\gmain$ and $\mhatmain$ and hence $\inner{\gmain}{\gtil} = \inner{\mhatmain}{\gtil} = 0$.
%
Since $\inner{\gmain}{\gtil} = 0$, \zcref{eq:firstorder} simplifies to $\Delta \Lmain \approx -\alpha \inner{\gmain}{\Pmain \mhatmain}$, exactly the first-order change the main task alone would produce on this batch.

\begin{table}[t]
  \caption{Test-set main loss and spread of $R$ across 10 members on \gls{gl:H2CM}, as mean and standard deviation over three seeds.
  A good method keeps the fit unchanged while revealing disagreement in the unobserved partitioning.}
  \label{tab:loss_vs_spread}
  \centering
  \begin{tabular}{lcccc}
    \toprule
    Method & median $\Lmain^{\mathrm{test}}$ $\downarrow$ & max $\Lmain^{\mathrm{test}}$ $\downarrow$ & $R$ std $\uparrow$ & $R$ spread $\uparrow$ \\
    \midrule 
    Baseline & $1.919 \pm 0.017$ & $2.130 \pm 0.010$ & $0.009 \pm 0.003$ & $0.031 \pm 0.013$ \\
    \gls{gl:PCGrad} & $1.887 \pm 0.025$ & $2.163 \pm 0.026$ & $0.177 \pm 0.007$ & $0.621 \pm 0.009$ \\
    \gls{gl:Bloop} & $2.135 \pm 0.064$ & $2.429 \pm 0.152$ & $0.103 \pm 0.015$ & $0.340 \pm 0.044$ \\
    \gls{gl:NANO} (ours) & $1.876 \pm 0.016$ & $2.034 \pm 0.076$ & $0.143 \pm 0.026$ & $0.491 \pm 0.093$ \\
    \bottomrule
  \end{tabular}
\end{table}
 
\paragraph{Bounding the magnitude.}
Orthogonality fixes the direction of the auxiliary term but not its size, while the second-order term grows with $\|\gtil\|$.
Two rescalings bound it and realise \zcref{update:so-bounded}. 
With $\V{u} = \Pmain \mhatmain$ being the main step direction, the first bound is in $\ell_2$ norm, \ie
\begin{align}
\gtil \leftarrow \min\!\left(1, \frac{\|\V{u}\|}{\|\gtil\|}\right)\gtil ,
\label{eq:cap}
\end{align}
so the auxiliary term never dominates the step in raw length.
The second bound is measured in the geometry the preconditioner induces, \ie $\Pmain = \M{H}^{-1}$ with \(\M{H} = \operatorname{diag}(\sqrt{\vhatmain})\), the diagonal curvature proxy Adam's step implicitly inverts.
Substituting $\M{H}$ for the Hessian, a step $-\alpha\V{d}$ along \(\V{d} = \V{u} + \lambda\gtil\) contributes the second-order term \(\tfrac{1}{2}\alpha^{2}\, \V{d}^{\top} \M{H} \V{d}\).
Writing $\qmain = \V{u}^{\top} \M{H} \V{u}$ and $\qaux = \gtil^{\top} \M{H} \gtil$, we require the auxiliary contribution not to exceed the main one, \ie $\lambda^{2}\qaux \le \qmain$.
The largest $\lambda \in [0,1]$ satisfying this is obtained by
\begin{align}
\lambda = \min\!\Big(1,\, \sqrt{\qmain / \qaux}\Big), \qquad\text{resulting in}\qquad \V{d}^{\top} \M{H} \V{d} \le 2\,\qmain.
\label{eq:damping}
\end{align}
The cross-term \(\V{u}^{\top} \M{H} (\lambda\gtil)\) vanishes, since $\cmain = 0$ gives $\M{H}\V{u} = \mhatmain$ and $\gtil$ is orthogonal to $\mhatmain$. 
The second-order term of the step is therefore at most twice the one the main branch would have produced alone.
Both rescalings are non-negative scalars, so $\inner{\gmain}{\lambda \gtil} = \inner{\mhatmain}{\lambda \gtil} = 0$ continues to hold exactly. 
With decoupled weight decay $\gamma$, the step is \(\V{\theta}_{t+1} = (1 - \alpha \gamma)\, \V{\theta}_{t} - \alpha \V{d}\).
We call the resulting optimiser \gls{gl:NANO}, summarised in \zcref{alg:main}.
The decay term is identical to the one present in AdamW \citep{loshchilov2019} and does not interact with the projection.

\section{Experiments}
\label{sec:experiments}
\begin{figure}[t]
  \centering
  \definecolor{baselineblue}{HTML}{3B6EA5}
\definecolor{experimentorange}{HTML}{9E2A2B}
\definecolor{neutraldark}{HTML}{404040}

\def\violindir{./figures/violin}

\newcommand{\violinstats}[6]{%
  \draw[neutraldark, line width=0.4pt] (axis cs:#1,#2) -- (axis cs:#1,#6);
  \draw[neutraldark, line width=1.6pt, line cap=round] (axis cs:#1,#3) -- (axis cs:#1,#5);
  \fill[white] (axis cs:#1,#4) circle[radius=0.9pt];
}

\begin{tikzpicture}
\begin{axis}[
  width=0.7\linewidth, height=3.6cm, scale only axis,
  ylabel={Testing loss\\(\% from baseline median)},
  ylabel style={align=center, font=\small, anchor=south},
  xtick={1,2,3,4},
  xticklabels={Baseline, \gls{gl:PCGrad}, \gls{gl:Bloop}, \gls{gl:NANO} (ours)},
  xticklabel style={font=\small},
  yticklabel style={font=\small},
  xmin=0.4, xmax=4.6, ymin=-11, ymax=36,
  yticklabel={\pgfmathprintnumber{\tick}\,\%},
  ytick distance=5,
  ymajorgrids=true, grid style={gray!25},
  table/col sep=comma,
  violin/.style={fill=#1, fill opacity=0.3, draw=neutraldark, line width=0.7pt},
]

\addplot[violin=baselineblue]     table[x expr=1+\thisrow{dx}, y=y]{\violindir/baseline.csv};
\addplot[violin=experimentorange] table[x expr=2+\thisrow{dx}, y=y]{\violindir/pcgrad.csv};
\addplot[violin=experimentorange] table[x expr=3+\thisrow{dx}, y=y]{\violindir/bloop.csv};
\addplot[violin=experimentorange] table[x expr=4+\thisrow{dx}, y=y]{\violindir/nano.csv};

\input{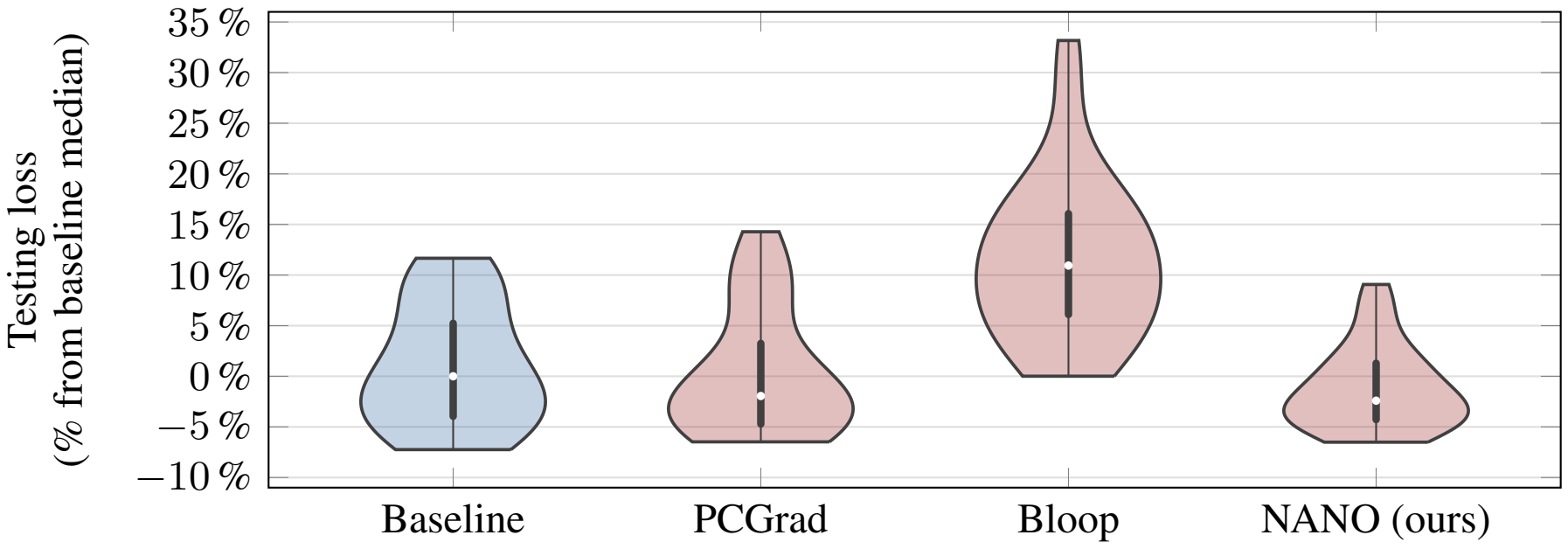}

\end{axis}
\end{tikzpicture}
  \caption{Main test loss of 10 members per method over 3 seeds, relative to the baseline median.
  A method preserves the fit if none of its members exceeds the baseline's worst test loss.}
  \label{fig:violin_plot}
\end{figure}

We train 10 \gls{gl:H2CM} members concurrently on identical batch sequences at \qty{1}{\degree} spatial and daily temporal resolution, repelled along the spatio-temporal mean transpiration $T$, and repeat each ensemble over three seeds.
We report the main loss $\Lmain^{\mathrm{test}}$ on the test set, and the diversity through the standard deviation and spread of the transpiration fraction $R=T/\mathrm{ET}$ over members.
As a baseline, we train a random-seed ensemble without $\Laux$.
\Gls{gl:NANO}, \gls{gl:PCGrad} \citep{yu2020} and \gls{gl:Bloop} \citep{hsieh2024} are all trained with the same repulsion loss.
\zcref{tab:loss_vs_spread} and \zcref{fig:violin_plot} report the spread in $R$ together with the main test loss relative to the baseline median.
Spread counts as equifinality only if the members producing it fit the observations as well as the baseline does.
\Gls{gl:NANO} raises the standard deviation of $R$ from $0.009$ to $0.143$ and its spread from $0.031$ to $0.491$, with all members below the baseline's worst test loss.
\Gls{gl:Bloop} yields a smaller spread at a worse fit.
\Gls{gl:PCGrad} attains the largest spread at a comparable median, but its worst members lie at or above the baseline's worst.

\section{Limitations}
\label{sec:limitations}
First, neutrality is approximate. 
Orthogonality and the second-order bound hold only per step, so residuals can accumulate into a drift in $\Lmain$. 
Moreover, $\M{H} = \operatorname{diag}(\sqrt{\vhatmain})$ is a running gradient magnitude, not a curvature, so the cross term need not vanish under the true Hessian.
Second, the repulsion acts only between nearest neighbours, so the spread we obtain is a lower bound on the equifinality present.
Third, the scheme costs one extra backward pass per member per step, and since $\Laux$ depends on the neighbours' current $\mu$, the $M$ members advance in lockstep, paced by the slowest.
Fourth, we neither ablate \gls{gl:NANO}'s components nor sweep the repulsion strength per method.
Finally, we test one model and one diagnostic. 
The update uses only the two gradients and AdamW's moments, and assumes nothing about the model, the loss, or $\mu$ beyond differentiability, so we expect it to transfer. 
Only $\tau$, which depends on $\mu$, must be set anew for each diagnostic.

\section{Conclusion}
We elicit the equifinality in \gls{gl:H2CM} by repelling ensemble members along the unobserved transpiration, subject to the fit to observations not degrading.
The constraint has to be imposed inside the optimiser: \gls{gl:NANO} accumulates both AdamW moments from the main gradient alone, projects the preconditioned auxiliary direction, dynamically bounds its length and steers the repulsion towards the coordinates the main gradient leaves idle.
The main objective then changes to first order exactly as under AdamW alone, with a second-order term at most twice that of AdamW in the preconditioner's geometry.
On \gls{gl:H2CM}, this raises the spread in $R$ from $0.031$ to $0.491$, with all members below the baseline's worst test loss.


\printbibliography


\newpage
\appendix

\section{Main loss}\label{app:h2cm_loss}
We use the main loss of \gls{gl:H2CM}~\citep{baghirov2026},
\begin{equation}
  \Lmain = \sum_{k=1}^{K} \frac{1}{N_k} \sum_{\ell=1}^{N_k} \bigl( y_{k,\ell} - \hat{y}_{k,\ell} \bigr)^2 ,
  \label{eq:main-loss}
\end{equation}
where $y_{k,\ell}$ and $\hat{y}_{k,\ell}$ are the observed and predicted values of data constraint $k$, both $z$-standardised per constraint, and $N_k$ is the number of data points of constraint $k$.
The constraints are monthly terrestrial water storage anomalies, snow water equivalent, evapotranspiration, runoff, fraction of absorbed photosynthetically active radiation, gross primary productivity and net ecosystem exchange, together with the mean annual carbon use efficiency.

\section{Algorithm}
\begin{algorithm}[]
\caption{One optimisation step for member $m$}
\label{alg:main}
\begin{algorithmic}[1]
\State $\gmain \gets \nabla_{\V{\theta}}\Lmain$; \quad $\gaux \gets \nabla_{\V{\theta}}\Laux$
\State update $\mmain, \vmain$ by \zcref{eq:moments}, bias-correct to $\mhatmain, \vhatmain$
\State build $\Pmain, \Paux$ by \zcref{eq:precond} with floors $\cmain \le \caux$ \Comment{\zcref{update:aux-biased}}
\State $\V{u} \gets \Pmain\,\mhatmain$; \quad \(\M{H} \gets \operatorname{diag}\big(\sqrt{\vhatmain}\big)\)
\State \(\gtil \gets \Pi\big(\Paux\,\gaux;\, \gmain,\, \mhatmain\big)\) by \zcref{eq:projpost} \Comment{\zcref{update:fo-descent}}
\State \(\gtil \gets \min\!\big(1, \|\V{u}\|/\|\gtil\|\big)\,\gtil\) \Comment{\zcref{update:so-bounded}}
\State $\qmain \gets \V{u}^{\top} \M{H} \V{u}$; \quad $\qaux \gets \gtil^{\top} \M{H} \gtil$
\State \(\lambda \gets \min\!\big(1, \sqrt{\qmain/\qaux}\big)\) \Comment{\zcref{update:so-bounded}}
\State \(\V{\theta} \gets (1-\alpha\gamma)\,\V{\theta} - \alpha\,(\V{u} + \lambda \gtil)\)
\end{algorithmic}
\end{algorithm}

\section{Hyperparameters}
\label{appendix:hyperparams}
All training settings are inherited from the \gls{gl:H2CM} baseline or are optimiser defaults: AdamW with base learning rate $\num{5e-3}$, linear warmup over 25 epochs from a tenth of that value, cosine annealing to $\num{5e-5}$ over the remaining 275, batch size 512, gradient clipping at value $0.5$, and moment decay rates $\beta_{1} = 0.9$ and $\beta_{2} = 0.999$.

The repulsion \zcref{eq:repulsion} uses $\varepsilon_{\mu} = \num{e-6}$ and $\tau = 0.1$.
Since $\tau$ depends on $\mu$, it sets the distance in mean transpiration below which neighbouring members repel.
$\tau = 0.1$ was chosen as a separation sufficient for the members to be distinct.
The preconditioner floors of \zcref{eq:precond} are $\cmain = 0$, leaving the main branch exactly AdamW's, and $\caux = 0.1$, chosen empirically, with $\varepsilon_{0} = \num{e-8}$.
The second projection direction in \zcref{eq:projpost} is dropped when the component of $\mhatmain$ orthogonal to $\gmain$ falls below $\num{e-3}\|\mhatmain\|$, where it is numerically unreliable.

For the comparison experiments, all three methods share the same $\Laux$.
\gls{gl:PCGrad} is the one-sided variant described in \zcref{sec:method}, whose combined gradient $\gmain + \pi(\gaux;\gmain)$ is passed to AdamW.
\gls{gl:Bloop} uses $\rho = 0.01$ for its gradient EMA, the value used throughout the original work, and the auxiliary weight $w = 1.0$.

\section{Computational resources}
\label{appendix:compute}
All runs were performed on a cluster node with NVIDIA A100 GPUs, one per ensemble member, so an ensemble of $M = 10$ members occupies 10 GPUs and 160 CPU cores.
Training one ensemble for 300 epochs takes roughly \qty{12}{\hour} with our update, against \qty{9}{\hour} for the random-seed baseline, the difference being the extra backward pass per member per step.

\section{Per-member $\mathrm{ET}$ fit}

\begin{figure}[t]
  \centering
  \input{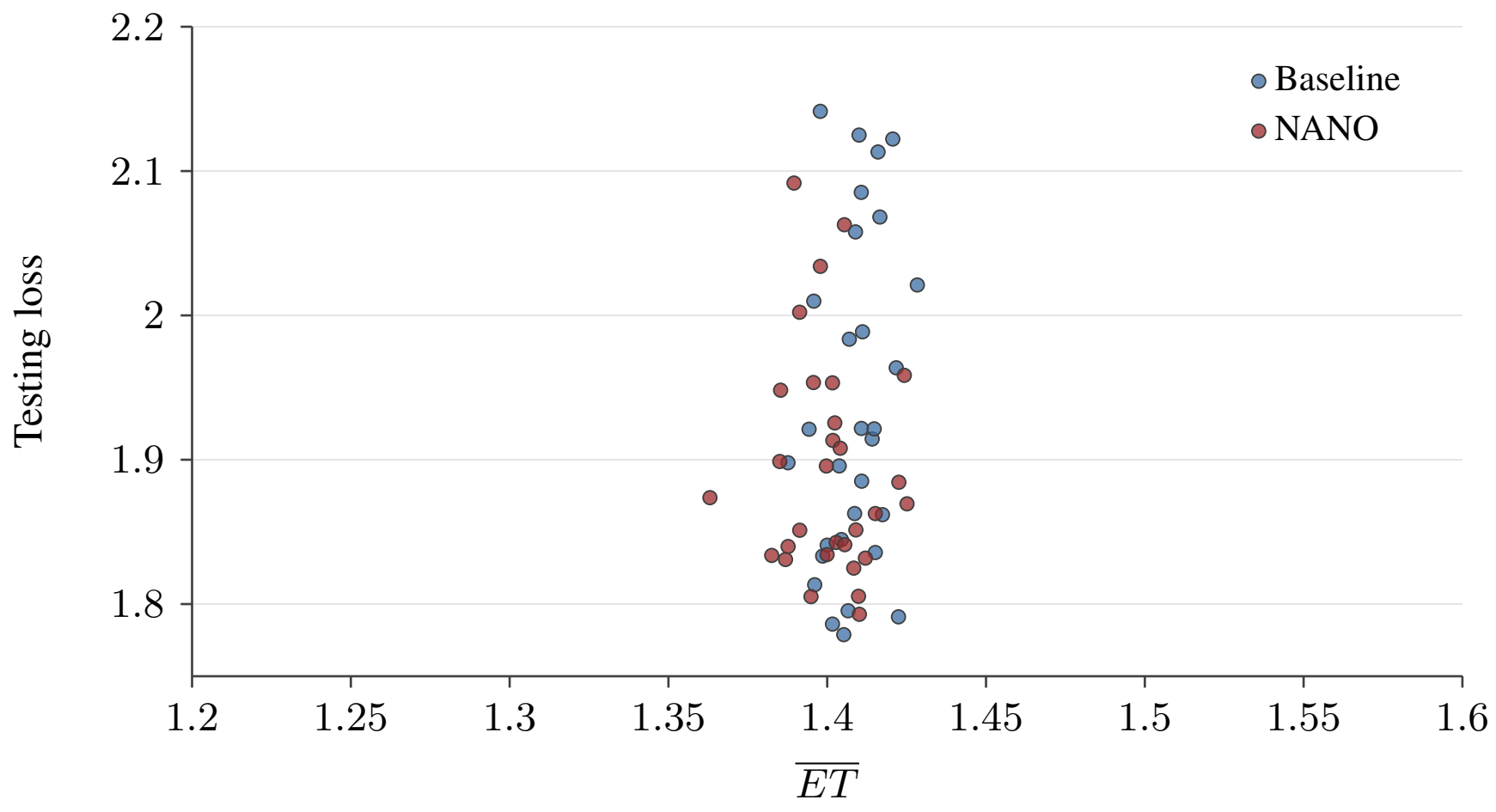}
  \caption{Per-member mean $\mathrm{ET}$ against the main test loss, for the baseline and \gls{gl:NANO} ensembles over three seeds.}
  \label{fig:et_fit}
\end{figure}

\zcref{tab:loss_vs_spread} and \zcref{fig:violin_plot} report the aggregate main loss of \zcref{eq:main-loss}, which could hide a shift in $\mathrm{ET}$ compensated by the other constraints. 
\zcref{fig:et_fit} therefore shows the mean $\mathrm{ET}$ of each member of the baseline and \gls{gl:NANO} ensembles over three seeds. 
The \gls{gl:NANO} members span a range of $\mathrm{ET}$ similar to the baseline's, so the repulsion redistributes $\mathrm{ET}$ between its components rather than changing total $\mathrm{ET}$.

\end{document}